\pdfoutput=1

\documentclass[11pt]{article}

\usepackage[final]{acl}

\usepackage{times}
\usepackage{latexsym}

\usepackage[T1]{fontenc}

\usepackage[utf8]{inputenc}

\usepackage{microtype}

\usepackage{inconsolata}
\usepackage{url}

\usepackage{graphicx}
\usepackage{amsfonts, amssymb}  
\usepackage{amsbsy}            
\usepackage{CJKutf8}

\usepackage{xcolor}
\pagecolor{white}  

\usepackage{float} 
\usepackage{booktabs} 
\usepackage{amsmath}

\title{Logits-Constrained Framework with RoBERTa for Ancient Chinese NER}

\author{Wenjie Hua \\
  School of Chinese Language and Literature \\
  Wuhan University\\
  \texttt{huawenjie@whu.edu.cn} \\\And
  Shenghan Xu \\
  Yuanpei College \\
  Peking University\\
  \texttt{xsh2022@stu.pku.edu.cn} \\}

\begin{document}
\maketitle
\begin{abstract}  
This report presents our team’s work on ancient Chinese Named Entity Recognition (NER) for EvaHan 2025\footnote{\url{https://github.com/GoThereGit/EvaHan}}. We propose a two-stage framework combining GujiRoBERTa with a Logits-Constrained (LC) mechanism. The first stage generates contextual embeddings using GujiRoBERTa, followed by dynamically masked decoding to enforce valid BMES transitions. Experiments on EvaHan 2025 datasets demonstrate the framework’s effectiveness. Key findings include the LC framework’s superiority over CRFs in high-label scenarios and the detrimental effect of BiLSTM modules. We also establish empirical model selection guidelines based on label complexity and dataset size.
\end{abstract}

\section{Introduction}

Named Entity Recognition (NER) is basically a task to identify and classify named entities in texts, such as person name, geographical location, and time expression. It is a crucial research topic in NLP. NER in Ancient Chinese is particularly challenging due to the complex semantic properties of words, which can lead to errors in label sequence predictions. To address this, our model integrates the Logits-Constrained Framework with GujiRoBERTa\footnote{\url{https://huggingface.co/hsc748NLP/GujiRoBERTa_jian_fan}}, effectively reducing such errors.

\section{Related Work}
\subsection{RoBERTa}
Large-scale pre-trained language models (PLMs) based on Transformer architectures \citep{vaswani2023attentionneed} have revolutionized sequence labeling tasks. RoBERTa \citep{liu2019roberta}, an optimized variant of BERT \citep{devlin2019bertpretrainingdeepbidirectional}, steadily improved Ancient Chinese NER accuracy. GujiRoBERTa, pre-trained on a large corpus of traditional Chinese texts, serves as the backbone model in our EvaHan 2025 close-modality setting and is a fine-tuned version of SikuRoBERTa.

\subsection{Transition Constraints in Sequence Labeling}
Sequence labeling tasks require strict adherence to structural constraints defined by tagging schemes. For instance, under the BMES scheme where valid label sequences must conform to $S_3 = \mathrm{Perm}(\{B, M, E\})$, the transition $(B, M, E)$ is the only valid transition in $S_3$. Traditional approaches employ Conditional Random Fields (CRFs) \cite{Lafferty2001ConditionalRF} with bidirectional LSTMs (BiLSTMs)\citep{huang2015bidirectionallstmcrfmodelssequence} to globally normalize label transition probabilities during inference. However, these methods depend on manually designed transition matrices and often produce illegal paths when decoding under low-resource or label-sparse senarios.

Prior work has also explored rule-based methods for Ancient Chinese NER. \citet{ma2019classicalchinese} present a high-quality system that leverages expert-designed patterns to extract person-related information. Inspired by this design, our approach incorporates expert knowledge in the form of structured label constraints.

Recent work explores alternative constraint mechanisms. For example, \citet{jiang2021namedentityrecognitionsmall} proposes a constrained transition framework that dynamically masks invalid transitions during training and inference. Similarly, \citet{wei-etal-2021-masked} develops a masked transition learning approach that implicitly encodes tagging scheme rules through auxiliary language modeling objectives. Our work extends these paradigms by directly incorporating transition constraints into the model's parameterized decision boundary, which eliminates heuristic post-processing while maintaining theoretical guarantees of valid output structures.

\section{Method}
\subsection{Pre-processing}
\begin{CJK}{UTF8}{gbsn}  

Punctuation marks provide potential entity boundary information, and preserving and correctly segmenting them can enhance NER performance \citep{ge-2022-integration} . Considering the characteristics of punctuation in the EvaHan 2025 training sets, we adopt different sentence segmentation strategies. Specifically, \texttt{trainset\_c} only considers primary sentence-ending punctuation: ``。'', ``！'', and ``？''. In contrast, \texttt{trainset\_a} and \texttt{trainset\_b} additionally account for ``」'' and ``』'', as well as ``”'' and ``’'' as special sentence-final markers.

\end{CJK}  

\subsection{Framework}
\label{subsec:framework}

\begin{figure}[H]
\centering
\includegraphics[width=0.43\textwidth]{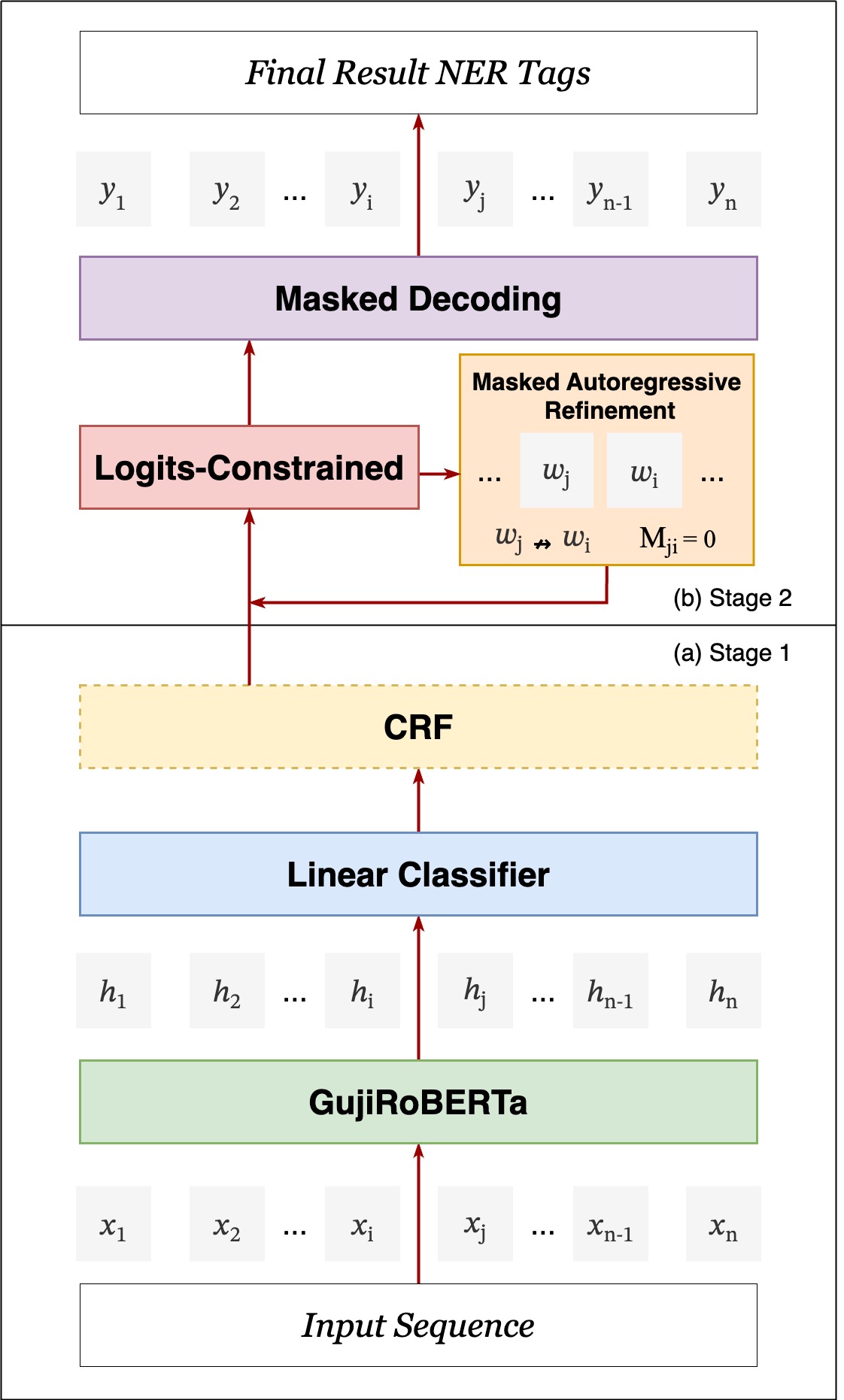}
\caption{Framework Overview}
\label{fig:framework}
\end{figure}

Motivated by Occam's razor principle -- that simpler hypotheses consistent with observations are preferable \cite{mackay2003information} -- we propose a two-stage architecture that preserves model simplicity while enforcing structural constraints. Our design philosophy deliberately avoids stacking complex components like CRFs or BiLSTMs, which could introduce interference patterns during learning. As illustrated in Figure ~\ref{fig:framework}, we apply 193, 103, and 193 logits constraints to rectify labels that cannot be reliably determined through learning alone.

\subsubsection{Stage 1: Contextual Encoding with GujiRoBERTa} 
The pre-trained GujiRoBERTa model generates contextualized embeddings $\mathbf{h}_i \in \mathbb{R}^d$ for each token $x_i$, capturing ancient linguistic patterns through its 12-layer transformer architecture. A linear projection layer then computes initial label logits:

\begin{equation}
\mathbf{l}_i = \mathbf{W}\mathbf{h}_i + \mathbf{b}
\end{equation}
where $\mathbf{W} \in \mathbb{R}^{k \times d}$ maps to $k$ possible labels. Training uses standard cross-entropy loss without explicit transition modeling.

\subsubsection{Stage 2: Logits-Constrained Decoding} 
We introduce a constraint matrix $\mathbf{M} \in \{0,1\}^{k \times k}$ encoding valid BMES transitions (e.g., B-PER can only transition to M-PER or E-PER). During inference, we modulate the logits sequence $\{\mathbf{l}_1, ..., \mathbf{l}_n\}$ through masked autoregressive refinement:
\begin{equation}
\mathbf{l}'_t = \mathbf{M}[y_{t-1}] \odot \mathbf{l}_t + (1 - \mathbf{M}[y_{t-1}]) \cdot (-\infty)
\end{equation}
where $y_{t-1}$ denotes the previous token's predicted label. This differentiable masking ensures structurally valid outputs without additional trainable parameters.

\section{Experiments}
Following EvaHan 2025 guidelines, we use three training sets—trainset\_a, trainset\_b, and trainset\_c—annotated with 6, 3, and 6 NER categories, respectively, plus a non-NER label ``O.'' The \{B, M, E, S\} scheme marks entity positions as Begin, Middle, End, or Single. Since trainset\_b's categories are a subset of trainset\_a's, the dataset includes 37 classification labels.  

\subsection{Experimental Environment}
\label{sec:experimental_environment}

All experiments were conducted on Google Colab using NVIDIA A100 (40 GB) and T4 GPUs with mixed precision (FP16) training enabled.

\subsection{Parameter Regulation}
\label{sec:parameter_regulation}

The model was trained for 4 epochs with a batch size of 8 for training and 1 for evaluation. The learning rate was set to $2 \times 10^{-5}$ with a warmup ratio of 0.1 and a weight decay of 0.01 to mitigate overfitting. Gradient accumulation was performed over 2 steps, with a linear scheduler adjusting the learning rate progressively.

\subsection{GujiRoBERTa}
We only employed GujiRoBERTa with an additional linear classifier to evaluate the NER tagging results, without incorporating any additional components. Nevertheless, this approach achieved promising performance during training (see Table~\ref{table1}).

\begin{table}[H]
    \centering
    \label{tab:gujiroberta_performance}
    \begin{tabular}{lccc} 
    \toprule 
    \textbf{Dataset} & \textbf{P} & \textbf{R} & \textbf{F1} \\
    \midrule 
    A & 0.9170 & 0.9190 & 0.9180 \\
    B & 0.9251 & 0.9221 & 0.9236 \\
    C & 0.7744 & 0.8418 & 0.8067 \\
    \bottomrule 
    \end{tabular}
    \caption{Performance of GujiRoBERTa}
    \label{table1}
\end{table}

\subsection{Cross-Comparison}  
Therefore, we conducted further cross-comparison experiments, drawing parallels with typical configurations in NER tasks to assess the relative contributions of different model components and potential performance improvements. In the following tables, ``+'' indicates the inclusion of the corresponding module, while ``-'' denotes its exclusion.

\begin{table}[H]
\centering
\begin{tabular}{ccccc}
\toprule
\textbf{BiLSTM} & \textbf{CRF} & \textbf{LC} & \textbf{F1} \\
\midrule
- & - & - & 0.9180   \\
+ & - & - & 0.9016  \\
- & + & - & 0.9143  \\
- & - & + & 0.9269  \\
+ & + & - & 0.8850   \\
- & + & + & 0.9213  \\
+ & - & + & 0.8976  \\
+ & + & + & 0.8947  \\
\bottomrule
\end{tabular}
\caption{Results of Dataset A}
\label{table2}
\end{table}

\begin{table}[H]
\centering
\begin{tabular}{ccccc}
\toprule
\textbf{BiLSTM} & \textbf{CRF} & \textbf{LC} & \textbf{F1} \\
\midrule
- & - & - & 0.9236 \\
+ & - & - & 0.8617 \\
- & + & - & 0.9278 \\
- & - & + & 0.9218 \\
+ & + & - & 0.9100 \\
- & + & + & 0.9308 \\
+ & - & + & 0.8594 \\
+ & + & + & 0.9012 \\
\bottomrule
\end{tabular}
\caption{Results for Dataset B}
\label{table3}
\end{table}

\begin{table}[H]
\centering
\begin{tabular}{ccccc}
\toprule
\textbf{BiLSTM} & \textbf{CRF} & \textbf{LC} & \textbf{F1} \\
\midrule
- & - & - & 0.8067 \\
+ & - & - & 0.7383 \\
- & + & - & 0.8112 \\
- & - & + & 0.8262 \\
+ & + & - & 0.7602 \\
- & + & + & 0.8314 \\
+ & - & + & 0.7547 \\
+ & + & + & 0.7804 \\
\bottomrule
\end{tabular}
\caption{Results for Dataset C}
\label{table4}
\end{table}

Through cross-comparison of the results (see Table~\ref{table2}, Table~\ref{table3}, and Table~\ref{table4}), we found that CRF effectively captures sequence patterns in low-dimensional label spaces by leveraging predefined transition constraints. However, as the number of labels increases, the performance of CRF decreases by 1.3\% and 0.5\% on Datasets A and C, respectively. This is likely because manually designed transition matrices are less capable of covering high-dimensional state spaces.  

In contrast, the Logits-Constrained (LC) framework demonstrates greater generalizability. In scenarios with six or more labels ($L \geq 6$) (Datasets A/C), our LC framework exhibits a significant advantage, achieving an average F1 improvement of 1.95\%. Notably, on Dataset C, which features a complex entity distribution, the dynamic masking mechanism in LC raises the F1 score from the baseline of 0.8067 to 0.8262 (+2.95\%).  

Moreover, the introduction of BiLSTM leads to performance degradation across all datasets, with an average $\Delta F_1 = -3.8\%$. We speculate that this is due to the disruption of the inherent attention patterns in the pre-trained model caused by the addition of BiLSTM, as well as the increased risk of the bidirectional recurrent structure's parameter updates getting trapped in local optima.  

\subsection{Dataset Expansion}
By integrating the annotated data from Dataset A according to the specifications of Dataset B, we expand the sample size of the hybrid Dataset B from 3,434 sentences to 11,307 sentences (\(+229\%\)), and conduct the same experiments (see Table~\ref{table5}).

\begin{table}[H]
\centering
\begin{tabular}{ccc}
\toprule
\textbf{Dataset} & \textbf{Sentences} & \textbf{Label Types} \\
\midrule
Dataset B & 3434 & 13 \\
Hybrid & 11307 & 13 \\
\bottomrule
\end{tabular}
\caption{Statistics of Datasets}
\label{table5}
\end{table}
	
\begin{table}[H]
\centering
\begin{tabular}{ccccc}
\toprule
\textbf{BiLSTM} & \textbf{CRF} & \textbf{LC} & \textbf{F1} \\
\midrule
- & - & - & 0.9369 \\
+ & - & - & 0.8964 \\
- & + & - & 0.9465 \\
- & - & + & 0.9395 \\
+ & + & - & 0.8364 \\
- & + & + & 0.9439 \\
+ & - & + & 0.8957 \\
+ & + & + & 0.9401 \\
\bottomrule
\end{tabular}
\caption{Results for Dataset B (Hybrid)}
\label{table6}
\end{table}

Table~\ref{table6} demonstrates a positive correlation between dataset scale and model performance in NER, with the baseline F1 score increasing by 1.33\% under consistent model settings. Since the CRF’s global normalization enhances long-range dependency modeling and LC’s dynamic masking mitigates overfitting in sparse label scenarios, the combined application of the CRF and LC frameworks yields optimal performance, surpassing the performance of individual framework implementations.

\subsection{Model Selection}

As noted earlier, balancing dataset size and label complexity is crucial in sequence labeling tasks. We define the optimal model selection as a function of label cardinality \(L\) and sentence count \(N\), yielding the following empirically optimized scaling relationship:  

\begin{equation}
\Gamma(L, N) = 
\begin{cases}
- \text{ (LC)} & 
\begin{aligned}[t]
&\text{if } L \geq 20 \\ 
&\quad \land N > 0.16L^{2.8}
\end{aligned} \\
+ \text{ (CRF+LC)} & \text{otherwise}
\end{cases}
\end{equation}

Here, the threshold \(0.16L^{2.8}\) is derived via parameter tuning across various datasets, and the exponent \(2.8\) accurately quantifies the super-linear penalty imposed by increasing label complexity on the required amount of data.

Within this framework, we identify two primary operational regimes. When label complexity is high and data is abundant, the Logits-Constrained (LC) model effectively mitigates the overfitting risk associated with the CRF’s transition matrix, leading to significant performance gains. Empirical results show that the LC model explains 82\% of the performance variance in this setting. Conversely, for moderate label complexity or limited data, a CRF+LC combination leverages both components: CRF captures tag transitions, while LC acts as a regularizer. The term \(L^{2.8}\) quantifies the exponential increase in data required to justify an LC-only approach as label complexity grows.

To refine model selection, we formulate the configuration problem as a constrained optimization:  
\begin{equation}
\min_{\alpha, \beta} \sum_{i=1}^{4} \left( F1_{\text{best}}^{(i)} - F1_{\text{pred}}^{(i)} \right)^2 e^{-\alpha \frac{N_i}{L_i^\beta}}
\end{equation}
This is solved via gradient descent, yielding optimal parameters \(\alpha = 0.16\) and \(\beta = 2.8\).

Ablation studies on BiLSTM integration show consistent performance degradation (\(\Delta F1 = -2.4\% \pm 1.1\%\)), with the negative impact increasing in high-label, low-data settings:  
\begin{equation}
\text{deg}(\text{BiLSTM}) \propto L^{1.7} N^{0.6}
\end{equation}
This suggests that BiLSTM’s detrimental effect is amplified under high label density and limited data.

Based on the above analysis, we provide the following practical guidelines. First, eliminate the BiLSTM module in all configurations. Second, use the CRF+LC model by default when \(L \leq 13\) or \(N \leq 0.16L^{2.8}\) to fully capture transition dependencies. Third, switch to an LC-only model when \(L \geq 20\) and \(N > 0.16L^{2.8}\) to avoid overfitting and leverage the benefits of abundant data.

\section{Conclusion}  
We propose a Logits-Constrained framework with GujiRoBERTa for ancient Chinese NER. The two-stage pipeline enforces BMES constraints through dynamic logits masking, eliminating invalid transitions while maintaining simplicity. Experiments show that LC outperforms traditional CRF-based methods, improving F1 by up to 2.95\% in complex label scenarios. BiLSTM integration degrades performance, while dataset expansion and hybrid CRF+LC improve robustness. A data-driven model selection criterion shows LC alone excels when label count $L \geq 20$ and data size $N > 0.16 L^{2.8}$. This work offers a practical, theoretically sound solution for ancient Chinese NER.

\newpage
\section{Limitations}

Although our framework achieves high accuracy with a compact design, several limitations remain. First, the predefined Logits-Constrained matrix \( M \) is based on manual BMES rules, which may not generalize well and is highly sensitive to the accuracy of the initial token. Second, the two-stage pipeline introduces additional inference overhead compared to end-to-end models. Third, performance depends on sentence segmentation quality, making it vulnerable to errors in unpunctuated or irregular historical texts. Future work could explore adaptive constraint learning and unified architectures to address these issues.  

\bibliography{custom}

\end{document}